\documentclass[conference]{IEEEtran}
\IEEEoverridecommandlockouts
\usepackage{cite}
\usepackage{amsmath,amssymb,amsfonts}
\usepackage{algorithmic}
\usepackage{graphicx}
\usepackage{textcomp}
\usepackage{xcolor}
\usepackage{listings}
\usepackage{hyperref}
\usepackage[normalem]{ulem}
\useunder{\uline}{\ulined}{}%
\DeclareUrlCommand{\bulurl}{}

\usepackage[numbers]{natbib}
\usepackage{hyperref}
\usepackage{cleveref}
\usepackage[htt]{hyphenat}
\def\BibTeX{{\rm B\kern-.05em{\sc i\kern-.025em b}\kern-.08em
    T\kern-.1667em\lower.7ex\hbox{E}\kern-.125emX}}
    
\lstdefinestyle{mystyle}{
    basicstyle=\ttfamily\footnotesize,
    breakatwhitespace=false,         
    breaklines=true,                 
    captionpos=b,                    
    keepspaces=false,                 
    numbers=right,                    
    numbersep=5pt,                  
    showspaces=false,                
    showstringspaces=false,
    showtabs=true,                  
    tabsize=3
}

\newcommand{\ttscale}[1]{{\fontsize{9}{10.2} \texttt{#1}}}

\begin{document}

\title{Semantic Modeling for Food Recommendation Explanations \\
\thanks{This work is partially supported by IBM Research AI through the AI Horizons Network.}
}
\author{\IEEEauthorblockN{1\textsuperscript{st} Ishita Padhiar}
\IEEEauthorblockA{\textit{Tetherless World Constellation} \\
\textit{Rensselaer Polytechnic Institute}\\
Troy, NY, USA \\
padhii@rpi.edu}
\and
\IEEEauthorblockN{2\textsuperscript{nd} Oshani Seneviratne}
\IEEEauthorblockA{\textit{Institute for Data Exploration and Applications} \\
\textit{Rensselaer Polytechnic Institute}\\
Troy, NY, USA \\
senevo@rpi.edu}
\and
\IEEEauthorblockN{3\textsuperscript{rd} Shruthi Chari}
\IEEEauthorblockA{\textit{Tetherless World Constellation} \\
\textit{Rensselaer Polytechnic Institute}\\
Troy, NY, USA \\
charis@rpi.edu}
\and

\IEEEauthorblockN{4\textsuperscript{th} Dan Gruen}
\IEEEauthorblockA{\textit{Institute for Data Exploration and Applications} \\
\textit{Rensselaer Polytechnic Institute}\\
Troy, NY, USA \\
gruend2@rpi.edu}
\and
\IEEEauthorblockN{5\textsuperscript{th} Deborah L. McGuinness}
\IEEEauthorblockA{\textit{Dept. of Computer Science, Tetherless World Constellation} \\
\textit{Rensselaer Polytechnic Institute}\\
Troy, NY, USA \\
dlm@cs.rpi.edu}
}

\maketitle
\begin{abstract}
With the increased use of AI methods to provide recommendations in the health, specifically in the food 
dietary recommendation space, there is also an increased need for explainability of those recommendations. 
Such explanations would benefit users of recommendation systems by empowering them with justifications for following the system's suggestions. 
We present the Food Explanation Ontology (FEO) that provides a formalism for modeling
explanations to users for food-related recommendations. FEO models food recommendations, using concepts from the explanation domain to create 
responses to user questions about food recommendations they receive from AI systems such as personalized knowledge base question answering systems. FEO uses a modular, extensible structure that lends itself to a variety of explanations while still preserving important semantic details to accurately represent explanations of food recommendations.
In order to evaluate this system, we used a set of competency questions derived from explanation types present in literature 
that are relevant to food recommendations. 
Our motivation with the use of FEO is to empower users to make decisions about their health, fully equipped with an understanding of the AI recommender systems 
as they relate to user questions, by providing reasoning behind their 
recommendations in the form of explanations. \\
Resource Website: \url{https://tetherless-world.github.io/food-explanation-ontology}\\
Ontology Link: \url{https://purl.org/heals/food-explanation-ontology/
}

\end{abstract}


\section{Introduction}
\label{sec:indroduction}
Food recommendation has become an essential method to help users adopt healthy dietary habits~\cite{tran2018overview}.
The task of computationally providing food and diet recommendations is challenging, as thousands of food items/ingredients have to be collected, combined in innovative ways, and reasoned over~\cite{freyne2011recipe}. 
Furthermore, there are many facets to the foods we consume, such as our ethnic identities, socio-demographic backgrounds, life-long preferences, all of which can inform our perspectives about 
the food we choose to consume to lead healthy lives. 
Food recommendation can get even more complicated when the food options available to an individual are further constrained because of a group setting 
(e.g., the seafood allergy of one family member may preclude recipes including shrimp to be recommended to the whole group). There is `no size fits all,' and even dietetic professionals have raised concerns that such varied dimensions need to be incorporated in their food recommendation advice~\cite{dietetics-white}. 
Such varied dimensions set up a need to provide food-related explanations to enhance the users' trust in recommendations made by food recommender systems, both automatic and human-driven, as users are more likely to follow the advice when the reasons for the advice are provided understandably.
Although explanations could help users trust in recommendations and encourage them to follow good eating habits, the inclusion of explanations into food recommender systems has not yet received the interest it deserves in the available literature~\cite{tran2018overview}.
Therefore, this work aims to bridge the gap between existing food recommendation systems by providing semantic modeling of explanations required in the complex and ever-expanding food and diet domain.

We introduce and discuss the FEO that extends the Explanation Ontology \cite{chari2020explanation} and the FoodKG (a food knowledge graph that uses a variety of food sources) \cite{haussmann2019foodkg} to model explanations in the food domain, a connection that is currently lacking in the current literature~\cite{tran2018overview}.
Our ontology can be classified under the post-hoc wing of Explainable Artificial Intelligence(XAI) and aims to interpret the results of black box AI recommender systems in a human-understandable manner \cite{adadi2018XAI, dosilovic2018XAI}. Accordingly, using a recommender system agnostic model, we aim to retroactively create  connections between the system and the recommendation,including modeling user details, such as allergies and likes, system details, such as location and time, and question details, such as parameters. 
We then assemble explanations by querying the ontology for different templates of knowledge types, defined using formalizations of explanation types. 
We add structure to the auxiliary modeling of the user, and the system, which we find are important components of comprehensive explanations to represent a range of explanation types \cite{chari2020explanation} (e.g., contextual, contrastive, counterfactual explanations) to model food-specific explanations, which would complement personalized, knowledge-based food recommendation applications such as the `Health Coach,' a healthy food recommendation service~\cite{RastogiSGCCHLSJ20}. 

\section{Related Work}
\label{sec:related-work}

Prior work has shown that users seek answers and reasoning for nutrition and food questions they might have
~\cite{forbes2011content,ueda2014recipe,mokdara2018personalized,ueda2011user,el2012food,Teng2012}. 
However, users are increasingly concerned with the evidence and reasoning that lead to those claims. 
Applying logic, reasoning, and querying on food and culinary arts have captured information, such as food categorization in FoodOn~\cite{dooley_foodon_2018}, recipes, and associated information in RecipeDB~\cite{recipedb}, and have brought together disparate sources of food information~\cite{popovski_foodontomap:_2019,haussmann2019foodkg}.
In the area of food recommendations, many existing approaches recommend recipes based on the recipe content (e.g., ingredients) ~\cite{el2012food,forbes2011content,Teng2012}, user behavior history (e.g., eating history)~\cite{forbes2011content,ueda2014recipe,mokdara2018personalized}, or dietary preferences~\cite{ueda2011user,mokdara2018personalized}. However, none of these systems provide the rationales for why the food was recommended, as these systems are utilizing black-box, deep learning models. Conversely, while there are systems that employ post hoc XAI methods to provide explanations for opaque AI systems \cite{adadi2018XAI, dosilovic2018XAI}, they have not yet been applied in the food recommendation domain.  
Our work differs from such previous works because we leverage a greater degree of explicit, semantic information about foods and other related semantically annotated data in generating explanations about recommending a food item or answering specific questions about a particular recommendation.
Additionally, the need to provide more user-centered explanations that help users improve trust and understanding of AI systems, and the information used for recommendations, has been gaining attention \cite{ mittelstadt2019explaining} lately. There have been some conceptual frameworks \cite{wang2019designing} and ontologies \cite{chari2020explanation} that attempt to model explanations from an end-user perspective. In our work, we aim to ground these more general-purpose efforts for the food domain. 

\citeauthor{dragoni2020explainable} describe a system based on logical reasoning that supports monitoring the users' behaviors and persuades them to follow healthy lifestyles, including recommending suitable food items, with natural language explanations\cite{dragoni2020explainable}. Their system performs reasoning to understand whether the users follow an unhealthy behavior regarding a food intake input.
Then the system generates the persuasion message with explanations using natural language templates~\cite{dragoni2020explainable}.
Our proposed, ontology-based method for generating explanations is complementary to their approach because we provide support for various types of explanations, not just trace-based explanations derived from templates for explaining the reasoner result.
We believe that by supporting different types of explanations, system developers will be supporting more user-friendly interfaces for personalized, consumer-facing applications \cite{chari2020explanation}. 
The primary aim for FEO is usage in 
more interactive or conversational food recommendations, for example, in a personalized health recommendation app.

\section{Ontology Modeling}
\label{sec:ontology-modeling}

Users typically have a diverse set of questions they might want to be answered beyond understanding the provenance behind a recommendation. Prior user studies in the clinical domain \cite{dan2020designing, wang2019designing} have indicated the need for diverse explanation types to address this range of questions. In this paper, we reuse this premise and build a semantic representation to model food explanations, borrowing from a previously identified set of user-centered explanation types \cite{chari2020explanation}. In the semantic representation, we cover the recommendations and knowledge components typically available to food recommender systems. 

\subsection{Ontology Structure }

We can view classes in our system from two different lenses. First, we want to be able to model them as components of explanations. Given a question, we have distinct object records, knowledge records, and recommendations that correspond to explanations captured in the Explanation Ontology \cite{chari2020explanation}. Additionally, the parameters within the users' questions, and the characteristics associated with these parameters, can be described in the context of food (the domain of application) and user and system semantics. To accurately model this interplay, we built up from each path to reach our final ontology structure.

To work in the space between these two domains, we 
started with base ontologies related to food and explanations. Specifically, the specifications for the explanations were borrowed from the Explanation Ontology (\url{http://purl.org/heals/eo}). The explanation ontology provides a semantic encoding for each explanation type, some of which we selected and expanded on for FEO's food domain. Further, we choose the  `What To Make' Ontology (\url{http://purl.org/heals/food}), which contains mappings to the more comprehensive FoodOn (\url{https://foodon.org}), for the food modeling foundation since it is concise that contains all relevant classes for typical food recommendation scenarios. In order to use a more standard ontology such as FoodOn, we would have to duplicate some of the classes and properties introduced in What to Make, such as \ttscale{User} and \ttscale{Ingredient}.
This ontology includes semantic models for food and recipes but required some expansion to provide comprehensive explanations for our food use case, which included adding modeling for different diets, as well as seasonal and regional availability.

In FEO, we attempt to model food explanations to provide answers to user questions that can be based on different types of knowledge, system recommendations, and user or system contexts. To be able to easily navigate between the different types of entities involved in explanations, we have introduced
two main concepts, the \textit{question parameter} and the \textit{ecosystem characteristics}, both of which are encapsulated under a \ttscale{feo:Characteristic} superclass.  Question parameters are typically entities of interest in a user's question, which can then be addressed via an appropriate explanation utilizing the \ttscale{feo:Parameter} class in the ontology. Ecosystem characteristics encompass various realms of the system, user, and knowledge, and we split them into \ttscale{feo:UserCharacteristic} and \ttscale{feo:System Characteristic} in the ontology. More specifically, the \ttscale{feo:Characteristic} class serves to map the question to the broader organization. 



\begin{figure}[!t]
\centerline{\includegraphics[width=1\columnwidth]{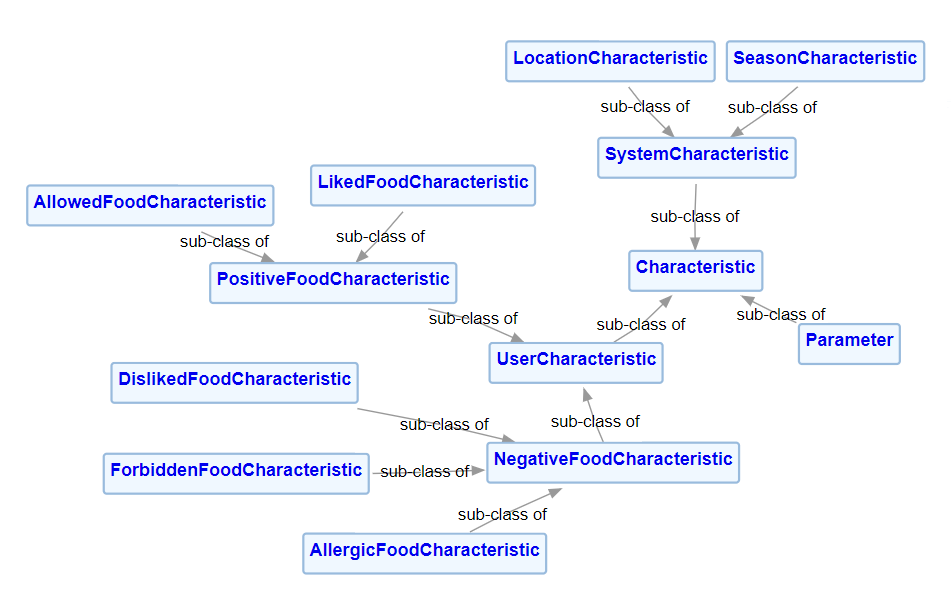}}
\caption{Subclasses of \ttscale{feo:Characteristic} class.}
\label{fig:characteristicModel}
\end{figure}

\begin{figure*}[!t]
\centerline{\includegraphics[width=1\textwidth]{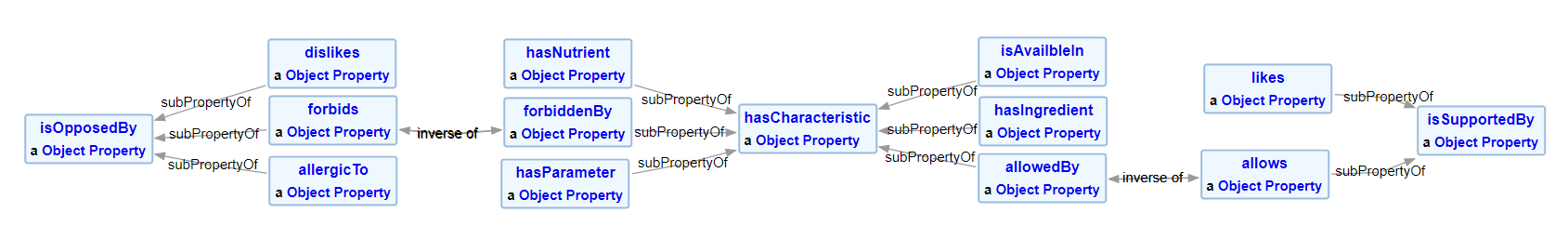}}
\caption{
Exemplar property relationships. 
The far left and right display  
two 
superproperties 
and the middle displays selected connections and interrelationships}
\label{fig:propertyModel}
\end{figure*}

Next, we discuss the three main sub-classes of \ttscale{feo:Characteristic}, which include \ttscale{feo:Parameter}, \ttscale{feo:UserCharacteristic}, and \ttscale{feo:SystemCharacteristic}. These classes describe details needed for explanations and are superclasses for most of the food domain-specific ontology classes. We use this hierarchy to organize and extract explanations from food characteristics. Our ontology uses inference to fit ecosystem instances by matching logical patterns for different characteristics.
For instance, the \ttscale{feo:LikedFoods} 
class corresponds to all the foods liked by a user, which we also classify as an \ttscale{owl:subClassOf} the \ttscale{feo:UserCharacteristic}. Similarly, we know that our system exists in a geographic region and season, enumerated as subclasses of \ttscale{feo:SystemCharacteristic}.
The last subclass of \ttscale{feo:Characteristic}, \ttscale{feo:Parameter}, performs a different form on the classification of data. The \ttscale{feo:Parameter} class is used to describe the different question inputs that we want to provide an explanation for. If our user asks the question: ``Why should I eat Food A?" we would classify Food A as a Parameter. 

The inclusion of the \ttscale{feo:Characteristic} superclass enables recursive querying on the ontology to get both matching and differing outputs. Given the same user and question, ``Why should I eat food A?" if we were looking for a contextual explanation, we would want to find all external knowledge that supports our domain (System, Goal, and User) attributes. To find these, we would iterate through all the characteristics of Food A, including Ingredient, Temporal, Location, or Diet information, and then look for characteristics of those attributes. Hence, rooting classes into the \ttscale{feo:Characteristic} hierarchy (\Cref{fig:characteristicModel}) allows us to recurse until we have found all the base characteristics of our question parameters and then check if they matched any of our environment characteristics that can be included within explanations. 

\subsection{Properties}

To appropriately connect the environment (system and knowledge stores) with the question posed by the user, we leveraged two essential restrictions on properties. 
 
\begin{figure}[!t]
\centerline{\includegraphics[width=.8\columnwidth]{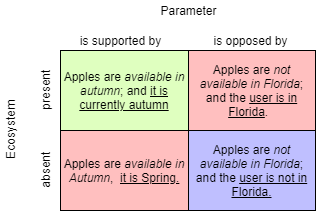}}
\caption{Defining facts and foils in our system. The italicized portion of each box refers to a base characteristic of the question \textit{parameter}. The underlined portion refers to the base characteristics of the \textit{ecosystem}. The characteristics themselves can overlap and can include things such as location and season. Instances in the green square would be considered facts, and those that are in the red boxes are foils. The blue box recommendation refers to instances that classify as neither facts nor foils.}
\label{fig:facts-foils}
\end{figure}

First, we used \ttscale{owl:subPropertyOf} classification and multiple inheritance to organize properties into relevant logical categories required for different explanation types. For instance, in trying to model Contrastive Explanations, it was important to distinguish between the facts that support one parameter and the foils that don't support another parameter. To find facts, we had to find all ``supportive" characteristics that match the environment and ``opposing" characteristics that don't match. We detail this intersection of parameters and the ecosystem in \Cref{fig:facts-foils}, which illustrates what is considered a fact versus a foil in the ontology.
Every positive characteristic of the parameter, which is also a positive characteristic of the system, is a fact. 
Every instance that supports a parameter and is absent in the ecosystem or opposes a parameter and is present in the ecosystem, is a foil. We formalize these relationships as equivalent classes to \ttscale{eo:Fact} and \ttscale{eo:Foil}, respectively.

In this case, we have \ttscale{feo:forbids} as an \ttscale{owl:subPropertyOf} of both \ttscale{feo:isOpposedBy} and \ttscale{feo:isCharacteristicOf}.
 
Next, we took advantage of the \ttscale{owl:inverseOf} property, which allows access to relationships between classes from both ends. In our case, this is necessary to use the reasoner to infer the \ttscale{feo:DislikedFoodCharacteristic}, which is made possible by defining the inverse property of \ttscale{feo:dislike}, and \ttscale{feo:dislikedBy}. Defining inverse properties helps enable inferences without having to explicitly specify specific facts about the user. 


Another key property we introduced to model the system was the \ttscale{feo:isInternal} data property. This boolean property is used to flag all the classes as either internal or external characteristics, where we define internal as being classes from the food and health domain. To illustrate, internal classes include \ttscale{Food}, \ttscale{Ingredient}, and \ttscale{Diet}, while external characteristics include \ttscale{Location}, \ttscale{Season}, and \ttscale{Budget}. These distinctions become increasingly important for the contextual explanation type, which, by virtue of their literature-derived and modified definition, only looks for explanations derived from external knowledge.

\begin{figure}[!t]
\centerline{\includegraphics[width=1\columnwidth]{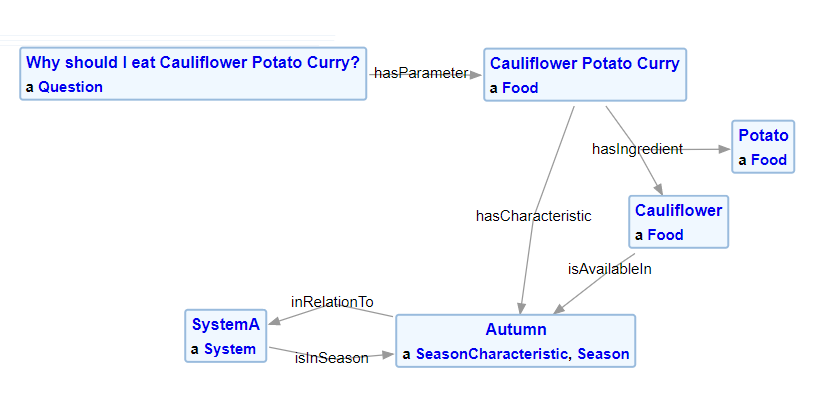}}
\caption{Subsection of the ontology that will demonstrate the class and property relationships discussed in Section \ref{sec:ontology-modeling}. 
}
\label{fig:q1Model}
\end{figure}

In \Cref{fig:q1Model}, we illustrate how these modeling choices work together to help us solve our competency questions. Specifically, it details how the ontology appears (after reasoning) with the classes necessary to answer the first competency question.

\section{SPARQL and Reasoning Process}
In order to extract explanations from FEO, we used a combination of logical reasoning and SPARQL queries. Much of the classification for this ontology relies on using a reasoner to infer class values and instances. Due to the large number of instances in the ontology, we use a reasoner known to handle individuals more efficiently, and we thus use the Pellet reasoner~\cite{sirin2007pellet}.  To meet our goal of providing food-specific explanations, we find that to provide comprehensive explanations, we require reasoning to be done on the ontology before we could query for explanations using the SPARQL query language. 

In order to generate explanations, our queries need to have access to the inferred characteristics. 
Therefore, we exported the ontology with the inferred axioms before running the SPARQL queries to evaluate the competency questions. We demonstrate the combination of such a reasoning and querying process in \Cref{lst:compQContext}, where a sample SPARQL query addresses competency question 1, providing a contextual explanation for the question ``Why should I eat Cauliflower Potato Curry?" To answer this question, given our query parameter, ``Cauliflower Potato Curry," we look to find all external characteristics (not food-related). Since we have defined the \ttscale{owl:hasCharacteristic} property to be transitive and defined its sub-properties as different food characteristics, once we run the reasoner, we can query for all types (properties) of different characteristics at all depths from the original ``Cauliflower Potato Curry" parameter. After this, we filter to find the \ttscale{feo:SystemCharacteristic}, 
which contains external values that match the characteristics of the parameter. In the results of the query, we can see \ttscale{feo:SeasonCharacteristic} with the value Autumn in the query results, which tells us that a contextual explanation for ``Cauliflower Potato Curry" is the current season in the system (i.e., ``autumn"). The season characteristic is defined as the current season for the region that the system is in, which is an explicitly defined relationship in the ontology.  

\section{Evaluation}
\label{sec:evaluation}

\begin{table}[!t]
\caption{Explanation Types and their corresponding food related example user questions. 
\label{tbl:explanationTable}}
\begin{center}
\newlength\q
\setlength\q{\dimexpr .42\columnwidth -2\tabcolsep}
\newlength\s
\setlength\s{\dimexpr .58 \columnwidth -2\tabcolsep}
\begin{tabular}{|p{\q}|p{\s}|}
\hline
\textbf{Explanation Types}& \textbf{Example Question} \\
\hline
Case-Based Explanations & What results from other users recommend food A?\\
\hline
Contextual Explanations & Why should I eat Food A?\\
\hline
Contrastive Explanations & Why was Food A recommended over Food B?\\
\hline
Counterfactual Explanations & What if we changed ingredient C? \\
\hline
Everyday Explanations & What foods go together? \\
\hline
Scientific Explanations & What literature recommends Food A? \\
\hline
Simulation-based Explanations & What if I ate food A everyday? \\
\hline
Statistical Explanations & What evidence from data suggests I follow diet D? \\
\hline
Trace-based Explanations & What steps led to recommendation E? \\
\hline
\end{tabular}
\end{center}
\end{table}

We employ a task-based evaluation \cite{raad2015evaluation} for our ontology using three main competency questions, each aimed at addressing a different explanation type that we attempted to extract from our model in FEO, as detailed in the following section. We have used competency questions as our method of evaluation as they are the accepted standard to ``evaluate the ontological commitments that have been made" \cite{gruniger1995evaluation}. 



As our model endeavors to provide explanations and context to users that get lost in black box AI models, we chose to evaluate the FEO by its ability to provide responses to a subset of important explanation types. In \Cref{tbl:explanationTable}, we have included a list of previously identified explanation types and the corresponding questions that might require an explanation for its food-related recommendation.
Post-hoc explanations provide an approximation of the rationales that users might be looking for \cite{dan2020designing}, which is what we wanted to tackle with the competency questions.


We support our choice in the selection of a subset of explanation types from Table \ref{tbl:explanationTable} for our evaluation via competency questions with observations from recent advances in the machine learning community, where we noticed that there is a focus on methods that generate contrastive and counterfactual explanations \cite{arya2019one}. Moreover, contrastive, counterfactual, and contextual explanations also contain explanations with scientific evidence, everyday evidence, and system trace. Therefore, an evaluation using these explanation types would also allow FEO to include other explanation types. Hence, we have completed our initial modeling to allow for \textbf{contrastive}, \textbf{counterfactual}, and \textbf{contextual} explanations and framed the evaluation of our ontology by these explanation types.

We undertook this process against the recommendations generated by the \emph{Health Coach} Application, which uses machine learning techniques to assess users' dietary needs and provide recommendations~\cite{RastogiSGCCHLSJ20}.

\subsection{Contextual Explanations}\label{sec:contextual_explanations}
A contextual explanation provides the user with any `external factors'  that affect the decision.
For example, in \Cref{lst:compQContext}, the season in which Cauliflower is available would be considered contextual knowledge that can be used to surface a contextual explanation, providing reasoning as to why Cauliflower Potato Curry should be eaten.
More generally, in the scope of this ontology and the food domain, we defined external factors to be characteristic of the user or system that are not directly food-related, such as seasonality, budget, and location.
    \begin{itemize}
        \item \textit{Health Coach-} The system recommends Cauliflower Potato Curry.
        \item \textit{Possible Question-} Why should I eat Cauliflower Potato Curry?
        \item \textit{Possible Answer-} Cauliflower Potato Curry uses the ingredient Cauliflower, which is available in the current season.
    \end{itemize}
\begin{lstlisting}[language=SPARQL,numbers=none, label=lst:compQContext]

SELECT DISTINCT ?characteristic ?classes
WHERE{
  ?WhyEatCauliflowerPotatoCurry feo:hasParameter ?parameter .
  ?parameter feo:hasCharacteristic ?characteristic .
  ?characteristic feo:isInternal False .
  ?systemChar a feo:SystemCharacteristic .
  ?userChar a feo:UserCharacteristic .
  Filter ( ?characteristic = ?systemChar || ?characteristic = ?userChar ) .
  ?characteristic a ?classes .
  ?classes rdfs:subClassOf feo:Characteristic .
  Filter Not Exists{?classes rdfs:subClassOf eo:knowledge }.
}
\end{lstlisting}

\begin{table}[!h]
\begin{center}
\begin{tabular}{|c|c|}
\hline
\textbf{?characteristic}& \textbf{?classes}  \\
\hline
\textit{feo:Autumn} & \ttscale{feo:SeasonCharacteristic}\\
\hline
\end{tabular}
\end{center}
\end{table}

\begin{lstlisting}[language=SPARQL,numbers=none, caption=SPARQL query and results for competency question 1. The query results indicate that the characteristics to explain why a user might want to eat Cauliflower Potato Curry., label=lst:compQContext]

\end{lstlisting}

\subsection{Contrastive Explanations}\label{sec:contrastive_explanations}

Contrastive explanations compare two different parameters. The parameters must be of the same type, and we look for \textit{facts} that support a question parameter and \textit{foils} that oppose the other. In \Cref{fig:facts-foils}, we define what facts and foils will look like in the ontology, which occurs at different intersections of the question parameters and the ecosystem characteristics. We want to return the facts that support only the first parameter and the foils that only oppose the second parameter. For instance, in the example question below, we can see that a relevant fact would be that Butternut Squash is available in autumn, which is the current season. Correspondingly, a relevant foil would be that Broccoli Cheddar Soup contains broccoli, which our user is allergic to.
    \begin{itemize}
        \item \textit{Health Coach-} Our user likes Broccoli Cheddar Soup. The system recommends Butternut Squash Soup.
        \item \textit{Possible Question-} Why should I eat Butternut Squash Soup over a Broccoli Cheddar Soup?
        \item \textit{Possible Answer-} Butternut Squash Soup is better than a Broccoli Cheddar Soup because Butternut Squash Soup is currently in season, and you are allergic to Broccoli Cheddar Soup.
    \end{itemize}
    
\begin{lstlisting}[language=SPARQL,numbers=none, label=lst:compQContrastive]

Select DISTINCT ?factType ?factA ?foilType ?foilB
Where{
  BIND (feo:WhyEatButternutSquashSoupOverBroccoliCheddarSoup as ?question) .
  ?question feo:hasPrimaryParameter ?parameterA .
  ?question feo:hasSecondaryParameter ?parameterB .

  ?parameterA feo:hasCharacteristic ?factA .
  ?factA a <https://purl.org/heals/eo#Fact>.
  ?factA a ?factType .
  ?factType (rdfs:subClassOf+) feo:Characteristic .
  Filter Not Exists{?factType rdfs:subClassOf <https://purl.org/heals/eo#knowledge> }.
  Filter Not Exists{?s rdfs:subClassOf ?factType}.
  
  ?parameterB feo:hasCharacteristic ?foilB .
  ?foilB a <https://purl.org/heals/eo#Foil> .
  ?foilB a ?foilType.
  ?foilType (rdfs:subClassOf+) feo:Characteristic .
  Filter Not Exists{?foilType rdfs:subClassOf <https://purl.org/heals/eo#knowledge> }.
  Filter Not Exists{?t rdfs:subClassOf ?foilType}.

}

\end{lstlisting}

\begin{table}[!h]
\begin{center}
\resizebox{\columnwidth}{!}{\begin{tabular}{|c|c|c|c|}
\hline
\textbf{?factType}& \textbf{?factA} & \textbf{?foilType}& \textbf{?foilB} \\
\hline
\texttt{feo:Season-} & \textit{feo:Autumn} & \texttt{feo:AllergicFood-} & \textit{feo:Broccoli}\\
\texttt{Characteristic} && \texttt{Charcteristic} &\\
\hline
\end{tabular}}

\end{center}
\end{table}

\begin{lstlisting}[language=SPARQL,numbers=none, caption=SPARQL query and results for competency question 2. The query results indicate that the characteristics to explain why a user might want to eat Butternut Squash Soup over a Broccoli Cheddar Soup., label=lst:compQContrastive]
\end{lstlisting}

\subsection{Counterfactual Explanations}\label{sec:counterfac_explanations}
Counterfactual explanations address the  \textit{What if} questions, attempting to answer if a change would occur if the system used question parameters for a recommendation. They provide an approximation of the effects of changing the inputs of the system. In this case, to provide an explanation, we would need to explore the effects of changing the factors in the user and system profile. This may include the user's nutritional goals, likes and dislikes, and certain restricting conditions like their pregnancy status, allergies, and diet. In the example below, we can see that sushi would be forbidden if the user were pregnant due to the fact that it contains raw fish. Further, in this example, the system has additional knowledge that foods high in folic acid are recommended for pregnancy, so we recommend Spinach Frittatas, which contain a high folate food (spinach). Also, suppose a dietary recommendation could be found for further evidence. In that case, such a recommendation could be encoded into the FoodKG and could also be included as additional reasoning or pregnancy alternatives into such a counterfactual explanation. 

    \begin{itemize}
        \item \textit{Health Coach-} The system recommends sushi.
        \item \textit{Possible Question-} What if I was pregnant?
        \item \textit{Possible Answer-}  If you were pregnant, you would be forbidden from eating sushi. You would be suggested to eat Spinach Frittata.
    \end{itemize}

\begin{lstlisting}[language=SPARQL,numbers=none, label=lst:compQcounterfact]

SELECT Distinct ?property ?baseFood ?inheritedFood
WHERE{
  feo:WhatIfIWasPregnant  feo:hasParameter ?parameter .
  ?parameter ?property  ?baseFood .
  ?property rdfs:subPropertyOf feo:isCharacteristicOf.
  ?baseFood a food:Food .
  OPTIONAL { ?baseFood feo:isIngredientOf ?inheritedFood.}
}

\end{lstlisting}

\begin{table}[!h]
\begin{center}
\begin{tabular}{|c|c|c|}
\hline
\textbf{?property}& \textbf{?baseFood} & \textbf{?inheritedFoods} \\
\hline
\texttt{feo:recommends} & \textit{feo:Spinach} & \textit{feo:SpinachFrittata} \\
\hline
\texttt{feo:forbids} & \textit{feo:Sushi} &\\
\hline
\end{tabular}
\end{center}
\end{table}

\begin{lstlisting}[language=SPARQL,numbers=none, caption=SPARQL query and results for competency question 3. The query results indicate that the characteristics to explain what might happen to food recommendations if the user became pregnant., label=lst:compQCounterfact]

\end{lstlisting}

\section{Future Work}
\label{sec:future-work}

We addressed three types of explanations (\textbf{contrastive}, \textbf{counterfactual}, and \textbf{contextual}) 
that would best serve the user's food-related questions. 
As case-based, everyday, simulation-based, and trace-based, scientific and statistical explanations have not yet been a focus of ML methods \cite{arya2019one}, we have deferred modeling them in this initial encoding of FEO.
Along those lines, the most intuitive direction in which to extend this work is adding functionality for more of the literature-derived explanation types we stated above, e.g., scientific and statistical explanations.

Specifically, this would include adding the modeling necessary to provide provenance-focused explanations, both scientific and statistical explanations.
For scientific explanations, we plan to use scientific knowledge from papers and studies as evidence that we share with the user as explanations. The difficulty with this process is in amassing enough knowledge to translate appropriately any evidence related question a user may match. Ideally, to provide the best evidence, we would find the knowledge that fits the user's characteristics and the question parameter. For a statistical explanation, we plan to use statistical knowledge to provide explanations. To accomplish this, we would want to aggregate data from the system that uses our ontology to find the users that accomplished similar goals as those of the user, with the same parameter as the user. The difficulties of this question include modeling for goal characteristics and determining what to search for to determine the values of the aggregate data.

After this, we want to integrate the FEO with larger datasets into a knowledge graph, allowing the user to query from a wider range of parameters. Currently, the ontology models the domain and would provide a well versed technical user with explanations. However, the ultimate goal of this work is to increase accessibility for the average user, which will be best served with more data. Along the same lines, we are integrating the ontology into a health application, `Health Coach'~\cite{RastogiSGCCHLSJ20}, that would be easier to use and more convenient for a non-technical user.

\section{Conclusion}
\label{sec:conclusion}
In this paper, we have discussed an ontology modeling for food and diet recommendation explanations, which aims to model and then be used to generate explanations specific to the context of the users and setting.
FEO is a domain-specific ontology where the domain concepts are abstracted up in a manner so that they can be comprehensively exposed to a user in the form of a diverse range of explanations. The class and property relationships that we detailed 
enable using simple queries to get explanations that explore many different variables. We strove to maintain the simplicity of the queries in order to ensure that a non-technical user can access explanations just as effectively as a technical user. From the modeling perspective, we found that the food domain would benefit from semantically bound explanations because of the variety of questions that a user might ask and the corresponding variety of explanations that they might require. From the user perspective, we chose the food domain specifically because food and diet are something that a growing number of people are concerned with, and we believe that our ontology can empower users to make informed decisions from their food choices. We plan to continue this work to extend the range of explanations that we can provide and increase accessibility to the tool by incorporating it into a more user-facing recommendation environment.

\bibliographystyle{IEEEtranN}    
{\fontsize{8}{10.2} \bibliography{references}}

\end{document}